\documentclass[letterpaper]{article} 
\usepackage{aaai24}  
\usepackage{times}  
\usepackage{helvet}  
\usepackage{courier}  
\usepackage[hyphens]{url}  
\usepackage{graphicx} 
\usepackage{natbib}  
\usepackage{caption} 
\usepackage{algorithm}
\usepackage{algorithmic}

\usepackage{soul}
\usepackage{amsmath}
\usepackage{mathtools}
\usepackage{subcaption}
\usepackage{textcomp}
\usepackage[toc,page]{appendix}

\usepackage{newfloat}
\usepackage{listings}
\DeclareCaptionStyle{ruled}{labelfont=normalfont,labelsep=colon,strut=off} 
\floatstyle{ruled}
\newfloat{listing}{tb}{lst}{}
\floatname{listing}{Listing}
\nocopyright
\title{ONNXExplainer: an ONNX Based Generic Framework to Explain Neural Networks Using Shapley Values}
\author{
    Yong Zhao\textsuperscript{\rm 1}\thanks{Corresponding author.}
    Runxin He\textsuperscript{\rm 2},
    Nicholas Kersting\textsuperscript{\rm 2},
    Can Liu\textsuperscript{\rm 1},
    Shubham Agrawal\textsuperscript{\rm 1},
    Chiranjeet Chetia\textsuperscript{\rm 1},
    Yu Gu\textsuperscript{\rm 2}
}
\affiliations{
    \textsuperscript{\rm 1}Visa Research, Austin, TX.\\
    \textsuperscript{\rm 2}Visa AI as Services, Austin, TX.\\
    yongz@visa.com
    
}

\usepackage{bibentry}

\begin{document}

\maketitle
  
\begin{abstract}
Understanding why a neural network model makes certain decisions can be as important as the inference performance. Various methods have been proposed to help practitioners explain the prediction of a neural network model, of which Shapley values are most popular. SHAP package is a leading implementation of Shapley values to explain neural networks implemented in TensorFlow or PyTorch but lacks cross-platform support, one-shot deployment and is highly inefficient. To address these problems, we present the ONNXExplainer, which is a generic framework to explain neural networks using Shapley values in the ONNX ecosystem. In ONNXExplainer, we develop its own automatic differentiation and optimization approach, which not only enables One-Shot Deployment of neural networks inference and explanations, but also significantly improves the efficiency to compute explanation with less memory consumption. For fair comparison purposes, we also implement the same optimization in TensorFlow and PyTorch and measure its performance against the current state of the art open-source counterpart, SHAP. Extensive benchmarks demonstrate that the proposed optimization approach improves the explanation latency of VGG19, ResNet50, DenseNet201, and EfficientNetB0 by as much as 500\%.
\end{abstract}

\section{Introduction}

Explainabile AI (XAI), one of the pillars of the greater Responsible AI programme coming of age,  is playing an increasingly vital role in deployments of machine learning models in commercial settings for two major reasons: first because such models are responsible for a growing portion of the automated decisions behind our everyday lives, where at the very least the consumers themselves want explanations of these decisions; and second because of the rising realization that most, if not all models, in particular Neural Networks (NNs) trained on a sufficiently large corpus, will contain hidden biases \cite{parrots} that one would like to expose and rectify wherever they drive a model decision, hopefully before impacting consumers. Not surprisingly, a large number of techniques have been proposed over the years to explain model decisions ( \cite{pdfs},  \cite{counterfactuals}, \cite{LIME}, \cite{lundberg2017unified}, \cite{permutations}, \cite{morris}, \cite{accumulated_local_effects}, \cite{anchors}, \cite{contrastive}, \cite{integrated_gradients}, \cite{GIRP}, \cite{protodash}, \cite{surrogates}, \cite{explainable_boosting}), of varying degrees of usefulness depending on the model and domain. Typically one analyzes model decisions after the fact (offline) with an arsenal of such techniques on an as-need basis, with SHAP \cite{lundberg2017unified} tending to hold an especially popular place for both its power, simplicity, and in some sense optimality as an explainer.

The production environment of large, sensitive industries such as consumer finance, however, necessitate immediate real-time responses to model decisions, ideally incorporating explainability into the model inference process,
yet XAI is not always an affordable option in leading frameworks. 
Open Neural Network Exchange (ONNX)~\cite{onnx} models deployed on a Triton~\cite{triton} inferencing framework, 
for example, enjoy superior performance~\cite{onnx_perform} to competing choices such as TorchServe \cite{torchserve} or Tensorflow Serving (TFS)~\cite{tfs}, yet, lacking a mechanism to store backwards layer derivatives (because all that matters for inference is forward-propagation!), SHAP/XAI support is absent.

We propose to remedy this situation with a low-latency explainer based on Shapley values~\cite{nowak1994shapley} specifically made for NN models, ONNXExplainer. Our contributions are summarized as following:
\begin{enumerate}
    \item We build a deployable automatic differentiation scheme in the ONNX ecosystem to obtain backpropagated gradients. This in turn provides the input data with a heatmap that shows the attributions of the features in modern inference servers (e.g., Triton).
    \item A novel optimization approach that pre-computes intermediate outputs for the reference samples substantially accelerates the explanations as seen in four benchmarked models in three frameworks (ONNX, TensorFlow, and PyTorch). In addition, our approach significantly reduces memory consumption when the number of reference samples is same as SHAP, which allows use of more reference samples and thereby enables better explanation with the same hardware constraints.
    \item One shot deployment, i.e. combining inference and explanations together as one model, is enabled to simplify on-boarding explainable models on device.
\end{enumerate}

\section{Related Work}
\label{sec:related_work}
As Machine Learning models are deployed in different industries and gradually playing an increasingly important role for stakeholders to make critical business decisions, the explainability of a model prediction becomes critically important.
Many methods have been proposed to explain machine learning models. In general, the model explainers can be divided into two categories: model-agnostic~\cite{LIME, strumbelj2010efficient, wachter2017counterfactual, mothilal2020explaining} and model-specific, such as deep learning explainers~\cite{shrikumar2017learning,GradCAM, binder2016layer,sundararajan2017axiomatic, simonyan2019deep, dhamdhere2018important}. In SHAP (SHapley Additive exPlanations)~\cite{lundberg2017unified}, the authors propose a unified framework to interpret machine learning models based on Shapley values~\cite{nowak1994shapley} and SHAP contains both model-agnostic and model-specific explainers.

Model-agnostic explainers, such as LIME~\cite{LIME} and kernel explanation in SHAP~\cite{lundberg2017unified}, perturbs the input and trains a surrogate model to approximate models' prediction to obtain explanations without opening up the black-box models. Among deep learning explainers, backpropagated (BP) gradients-based approaches predominate because it attributes the importance scores to each feature in the input naturally. Saliency maps have been used for some time to visualize/interpret the images~\cite{simonyan2013deep,dabkowski2017real}. Attribution propagation approach propagates the contributions of all neurons in the network to the input~\cite{sundararajan2017axiomatic, montavon2017explaining, shrikumar2017learning}. DeepLIFT~\cite{shrikumar2017learning} outperforms other BP methods by backpropagating negative relevance to increase class-sensitivity and solve the convergence problem and it is the only BP method that passes the test in one recent work which theoretically analyzes BP methods~\cite{sixt2020explanations}. 

In addition to LIME and SHAP, several other open source python packages for XAI have been introduced~\cite{dalex1,dalex2,omnixai,aix360,interpretml,captum}. Many of them are an unification of existing methods or mostly a wrapper of LIME and SHAP with interactive utilities. For instance, OmniXAI~\cite{omnixai} builds model-specific and model-agnostic explainers on top of LIME and SHAP. Captum~\cite{captum} is a different case with PyTorch interface which integrates several explainers for NNs. However, none of them allow framework interoperability or easy deployment on different hardware in modern inference servers. Our approach proposes to use DeepLIFT~\cite{shrikumar2017learning} to compute Shapley values for NNs in ONNX ecosystem, which allows us to combine inference and explanation together as one model file to deploy across various hardware in real time.

\section{Deep Learning Important FeaTures (DeepLIFT)}
\label{sec:deeplift}
In this manuscript we use an approximation to Shapley values  based on DeepLIFT, the philosophy of which is to explain the difference in output from some reference output in terms of difference of the input from the corresponding reference input, measuring the target input's importance on the model prediction through back-propagation~\cite{shrikumar2017learning}. 

In this section, we introduce mathematical operators and their notations in this manuscript.
For a neural network model, let $t$ denote the  output of a neuron in some intermediate layer and $x_0, x_1, \cdots, x_n$ denote the necessary and sufficient inputs to compute the $t$ from the neuron. 
The reference-from-difference $\Delta{t}$ is denoted as $\Delta{t}=t-t^0$, where $t^0$ is the corresponding output of the neuron from the reference input ${x^0}_0, {x^0}_1, \cdots, {x^0}_n$, which is chosen according to domain knowledge and heuristics (for MNIST digit recognition, for example, one could choose an all-black background image of 0's).
DeepLIFT assigns contribution scores $C_{\Delta{x_{i}}\Delta{t}}$ to $\Delta{x_{i}}$ s.t.,
\begin{equation}
    \sum^n_{i=1}C_{\Delta{x_{i}}\Delta{t}} = \Delta{t},
    \label{eq:s2d}
\end{equation}
where $C_{\Delta{x_{i}}\Delta{t}}$ is the amount of difference-from-reference in $t$, and it is attributed to or 'blamed' on the different-from-reference of $x_i$. 
The intuition is, as explained in \cite{lundberg2017unified}, that of a sort of fast approximation of Shapley values where we examine the effect of 'including' each $x_i$ in place of its reference default ${x^0}_i$.

Then the multiplier/derivative is defined as :
\begin{equation}
    m_{\Delta{x}\Delta{t}} = \frac{C_{\Delta{x}\Delta{t}}}{\Delta{x}},
    \label{eq:multiplier}
\end{equation}
where $\Delta{x}$ is the difference-from-reference in input $x$ and $\Delta{t}$ is the difference-from-reference in output $t$. 
Since the contribution of $\Delta{x}$ to $\Delta{t}$ is divided by the input difference, $\Delta{x}$, we can use the multiplier as a discrete version of a partial derivative~\cite{shrikumar2017learning}. 

The chain rule for the multiplier can then be defined as the following,
\begin{equation}
    m_{\Delta{x_i}\Delta{t}} = \sum^n_{j=1}m_{\Delta{x_i}\Delta{y_j}}m_{\Delta{y_j}\Delta{t}},
    \label{eq:chain_rule}
\end{equation}
where $x_i$ is the neuron input for layer $H_{l}$; $y_{0},y_{1},\cdots,y_n$ are neuron outputs for layer layer $H_{l}$ and neuron inputs for successor of $H_{l}$.
The analogy to partial derivatives allows us to compute the contributions of model output w.r.t. model input via back-propagation. 
The Shapley values can then be approximated by the average as:
 \begin{equation}
     \phi \approx Avg(\mathcal{M} * (X - R)),
 \end{equation}
where $\mathcal{M}$ is the final matrix computed by multiplier w.r.t. the model input in the back-propagation and $X$ is the input and $R$ is input's reference.
Our focus is how to implement and accelerate $\mathcal{M}$ in the ONNX ecosystem for any neural network. 
In ONNXExplainer, we adjust gradient (multiplier) computation for nonlinear operators (e.g., \textit{Sigmoid} and \textit{MaxPooling}) and use the original gradients computation for linear operators (e.g., \textit{MatMul} and \textit{Conv}).

\section{Method}
\label{sec:method}
\begin{figure}[ht]
\centering
\includegraphics[width=8cm]{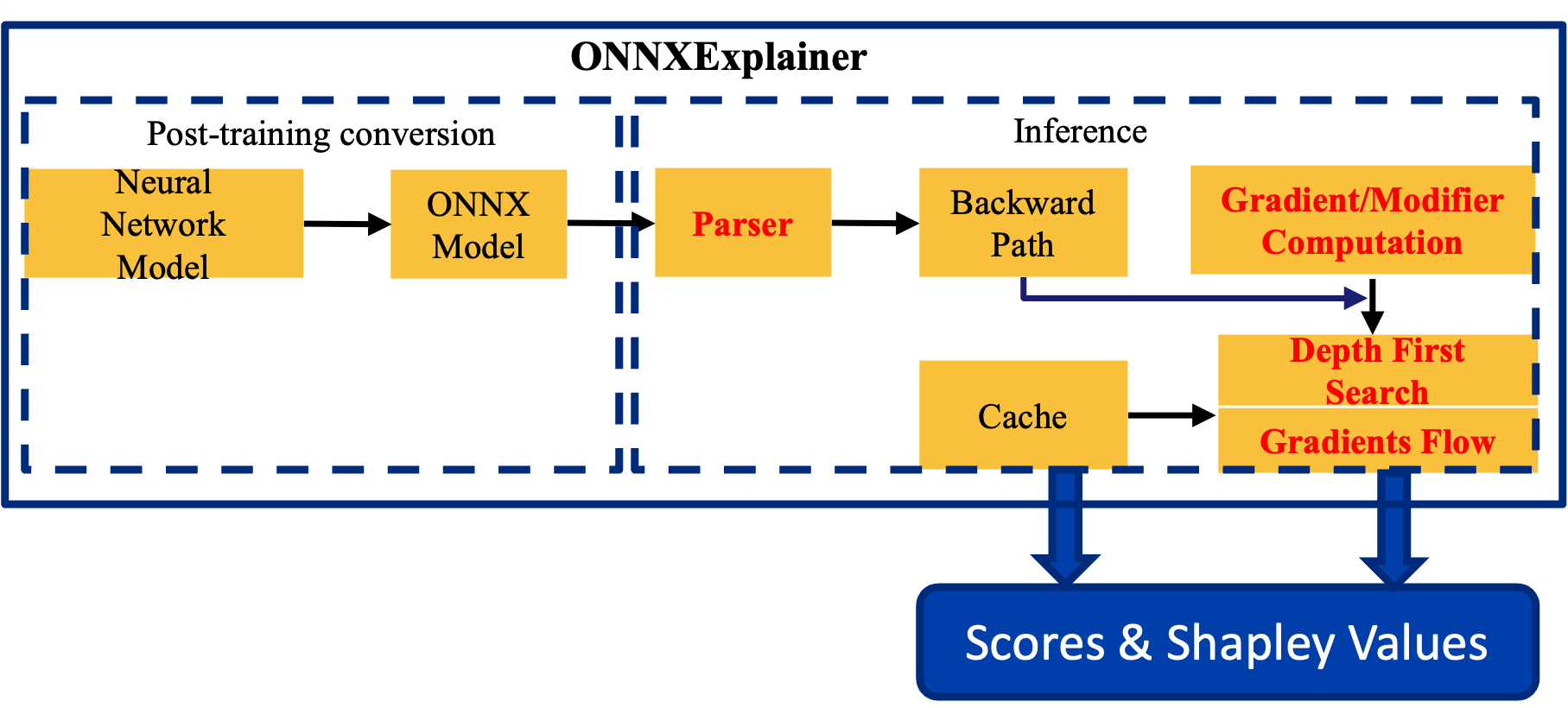}
\caption{A Schematic ONNXExplainer workflow for NN models, containing Parser, Gradient/Multiplier Computation, Automatic Differentiation (Depth First Search and gradient flows), and inference time optimization (Cache)}\label{fig:onnx_exp_dia}
\centering
\end{figure}

ONNX is an open-source standard  representing neural networks in order to make it easier for researchers and engineers to move NNs among different deep learning frameworks and computing devices~\cite{onnx}. 
We propose ONNXExplainer to explain and interpret NNs in the ONNX ecosystem. 
As in SHAP~\cite{lundberg2017unified}, ONNXExplainer uses DeepLIFT~\cite{shrikumar2017learning} to compute the Shapley values as a measure of feature importance for NNs. 
SHAP depends on the automatic differentiation mechanism in TensorFlow and PyTorch to compute gradients for Shapley value.
However, in onboarding deep learning models on a device for inference, current deep learning frameworks will discard gradient information and only keep the forward pass graph.
In this scenario, SHAP cannot be saved or called directly from the inference engine, e.g.,  ONNXruntime,  because of the missing dependencies and computation graphs to calculate differentiation and gradients.
Instead, ONNXExplainer provides its own automatic differentiation mechanism to compute gradients for NNs so that practitioners can use it to serve their NN models with Shapley values in a production pipeline. 
To make ONNXExplainer explain a general NN, it has three key components as shown in Figure~\ref{fig:onnx_exp_dia}: Neural Network Parser, Gradients/Multipliers Computation, and Automatic Differentiation. 
Moreover, an optimization approach is provided to improve the inference time by pre-computing intermediate outputs to forward propagation inside ONNXExplainer.

\paragraph{Neural Network Parser} As shown in Figure~\ref{fig:onnx_exp_dia}, a general NN model is converted to an ONNX format model. 
The ONNX model contains computation nodes to inference the inputs. After feeding the ONNX Model for NNs to ONNXExplainer, it first establishes the forward symbolic graph. In the forward symbolic graph one computation node is linked to other computation nodes because the output of the computation node is always either the input to other computation node(s) or the output of the ONNX model. 
With that, we can build the backward graph's main structure whose vertexes carry information about the computation nodes\footnote{The python style code for the Parser is provided in the Appendix A.}. 

\paragraph{Gradients/Multipliers Computation}
Each deep learning framework contains hundreds of operators. We need to implement/define gradients (for linear operators) and multipliers (for nonlinear operators) in the ONNX ecosystem as well. 
In the current manuscript, we have provided gradients/multipliers computation for more than 25 operators, including \textit{Concat}, \textit{Add}, \textit{Mul}, \textit{MatMul}, \textit{Gemm}, \textit{Sigmoid}, \textit{ReLU}, \textit{Softmax}, \textit{Conv}, \textit{MaxPool}, \textit{AveragePool}, \textit{GlobalAveragePool}, \textit{Transpose}, \textit{BatchNormalization}, and others. 
When executing the forward symbolic graph, some intermediate outputs needed for the gradient computation might be retained in memory for other operations~\cite{abadi2016tensorflow, paszke2017automatic, paszke2019pytorch}. 
In this way, the deep learning frameworks can avoid extra computations when training the NNs. However, those intermediate outputs usually are opaque to the users. 
After training, current deep learning frameworks in the market will freeze the model and keep only the forward symbolic graph for inference.
Under this scenario, we have to implement the gradients/multipliers computation for the inference operators only using the existing forward symbolic graph\footnote{Details are provided in the Appendix B.}. 
To summarize at a high level, we use the \textit{Linear Rule} for linear operations to compute gradients and the \textit{Rescale Rule} and the \textit{RevealCancel Rule} for nonlinear operations to compute multipliers~\cite{shrikumar2017learning}.

\paragraph{Automatic Differentiation}
Automatic differentiation is useful for implementing machine learning algorithms such as back-propagation for training neural networks. PyTorch and TensorFlow both use automatic differentiation. 
However, when the models get on-boarded on device for inference purpose, they usually are frozen as a file which containing only the model structure (forward pass) and corresponding parameters. 
DeepLIFT needs the back-propagation to compute Shapley values and SHAP is reliant on the \textit{tf.gradient} for \textit{TensorFlow} and the \textit{torch.autograd.grad} for \textit{PyTorch}, both of which are not serializable. Hence, we need to build our own automatic differentiation in the ONNX ecosystem. 

\begin{figure}[ht]
\centering
\includegraphics[width=8cm]{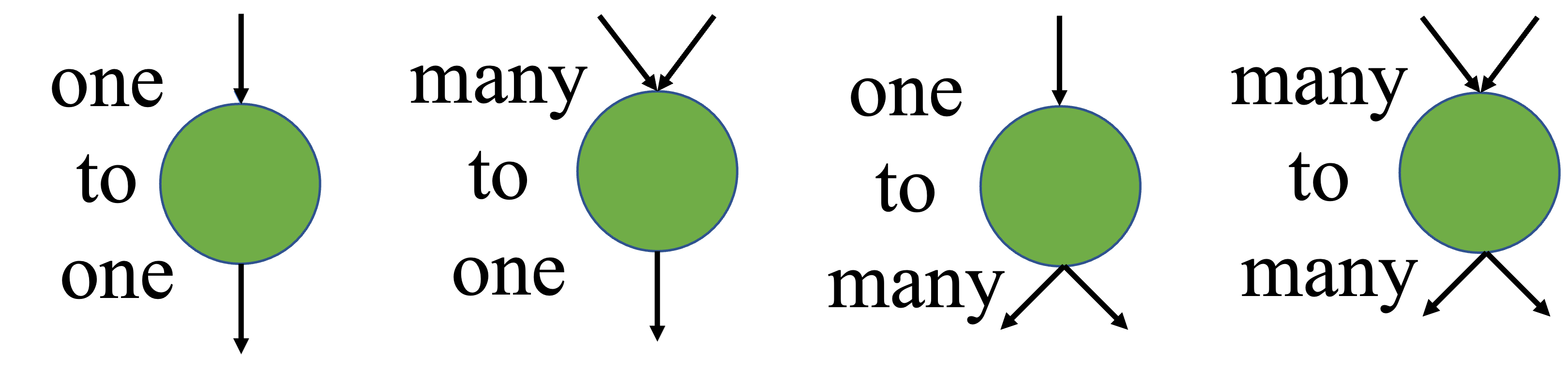}
\caption{Four types of gradient flows.}\label{fig:grad_flowin}
\end{figure}

The differentiation algorithm conducts Depth First Search (DFS) to identify all of the operators in the backward pass from the output to the input to the model and sums the partial gradients that each operator contributes. Before introducing how the DFS works, we demonstrate the four types of gradient flows as shown in Figure~\ref{fig:grad_flowin}:
\begin{itemize}
    \item one2one: The one2one type is simple: both incoming and outgoing gradients just have one branch. We multiply the incoming gradients with the local gradients if any to obtain the outgoing gradients. Activation functions are typical operators of  this type. If the operator has no local gradients, we just pass the incoming gradients to the successors in the backward pass.
    \item many2one: The many2one type has multiple flows of incoming gradients but one flow of outgoing gradients. We sum all incoming gradients at first and then multiply this summation with the local gradients if any to obtain the outgoing gradients.
    \item one2many: The one2many type has one flow of incoming gradients and multiple flows of outgoing gradients. After multiplying the incoming gradients with local gradients if any, we split or assign the outgoing gradients to the successors.
    \item many2many: The many2many type is the combination of many2one and one2many. 
\end{itemize}

\begin{algorithm}[tb]
\caption{DFS to compute Shapley values}
\label{alg:dfs}
\textbf{Input}: the backward graph $G$, the first computation node $N$  \\
\textbf{Output}: Gradient node list $L$
\begin{algorithmic}[1] 
\STATE Let $S$ be the stack.
\STATE $S$.push($N$)
\STATE Mark $N$ as visited.
\STATE Define the difference-from-reference $y_{x}-y_{r}$ as the loss $grad_{in}$. \COMMENT{$y$ is the output of model.}
\WHILE{$S$ is not empty}
\STATE $C$ $\gets$ $S$.pop()
\STATE $O, grad_{in} \gets F_{grad}(C, G, grad_{in})$ \COMMENT{$F_{grad}$ is the function to compute gradients/multipliers for operators.}
\STATE Append $O$ to $L$.
\FOR{neighbor $W$ of $C$ in $G$}
    \IF {$W$ is not visited and it gets all gradient flows}
    \STATE $S$.push($W$)
    \STATE Mark $W$ as visited.
    \ENDIF
\ENDFOR

\ENDWHILE
\STATE \textbf{return} $L$
\end{algorithmic}
\end{algorithm}

\begin{figure}[ht]
\centering
\includegraphics[width=.35\textwidth]{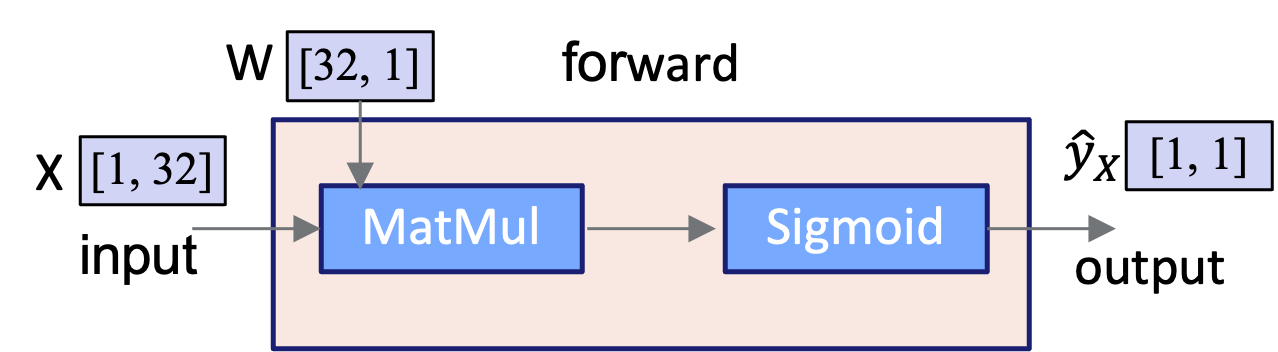}
\caption{Inference graph to the demo model.}\label{fig:demo_forward}
\end{figure}

\begin{figure}[ht]
\begin{subfigure}[t]{0.45\textwidth}
    \centering
    \includegraphics[width=.99\textwidth]{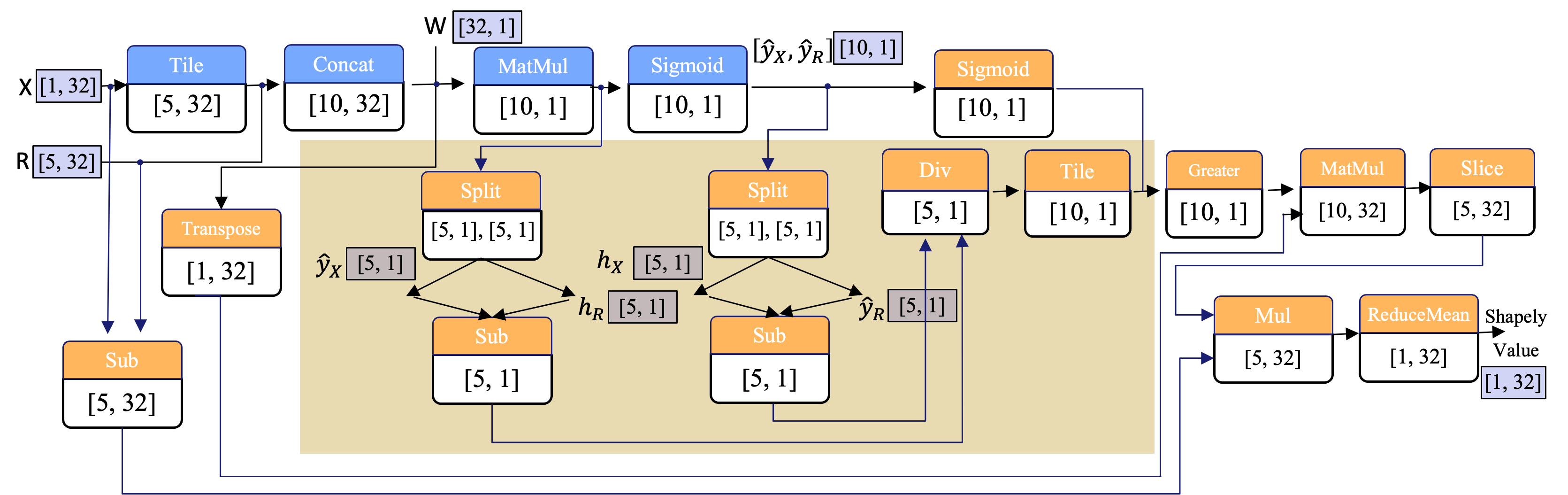}
    \caption{SHAP}
   \label{fig:shap_diagram}
\end{subfigure}%

\begin{subfigure}[t]{0.45\textwidth}
    \centering
    \includegraphics[width=.99\textwidth]{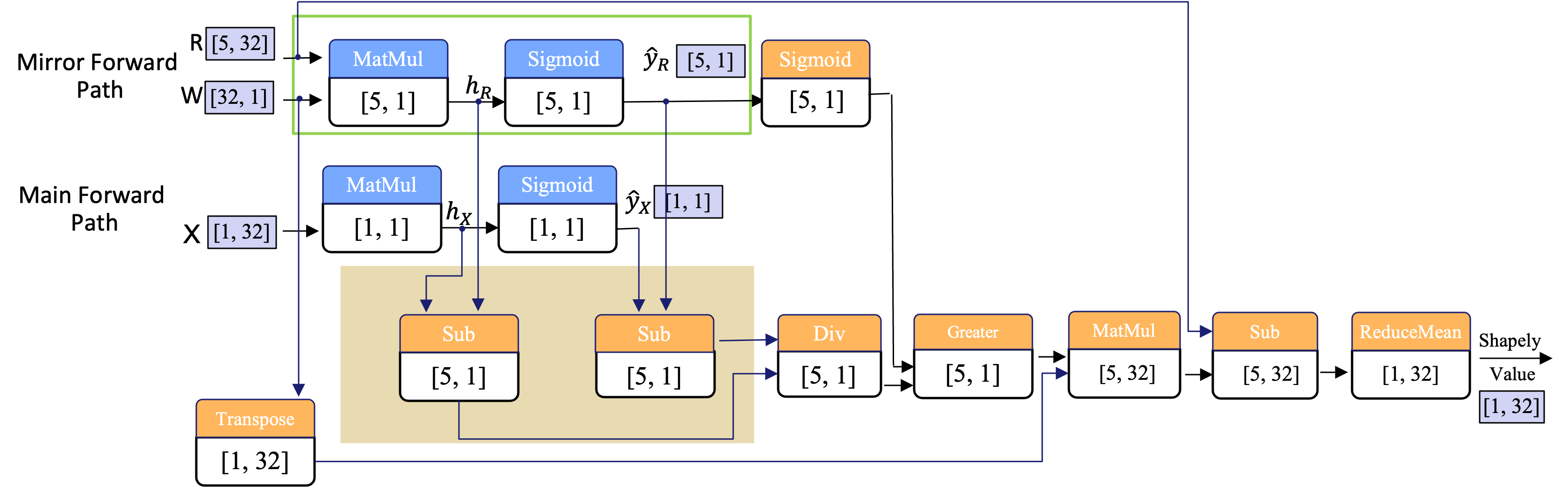}
    \caption{Optimization}
    \label{fig:opt_diagram}
\end{subfigure}%
\caption{The diagrams of computing Shapley values using SHAP and our optimization approach for a simple forward symbolic graph with two computation nodes: \textit{MatMul} and \textit{Sigmoid}. In diagrams, blue nodes are for the forward pass and orange nodes are for the backward pass. For instance, an orange Sigmoid node is the gradient computation of a blue Sigmoid node in the back-propagation. The grey area in each box is the output shape of the computation node.  We use 5 reference samples to explain the data in the example.}
\label{fig:opt_shap}
\centering
\end{figure}

ONNXExplainer uses DFS (Depth-First Search) to reverse the forward symbolic graph to compute Shapley values. The procedure is summarized in Algorithm~\ref{alg:dfs}. The DFS algorithm takes the backward graph $G$ and the first computation node $N$ as the inputs and it returns a list of computation nodes. Here the backward graph $G$ is obtained from Neural Network Parser and $N$ is the first computation node in the backward pass. Each vertex in $G$ contains information to perform DFS and we use the name of the visiting computation node to get that information. From lines 1-3, we create an empty stack and push $N$ onto the stack, marking $N$ as visited. Line 4 defines the loss $y_{x}-y_{r}$ to compute gradients w.r.t. the model input. The rest of the algorithm details how to traverse all computations nodes in the backward pass. Function $F_{grad}$ returns a list of computation nodes $O$ to compute gradients for the visiting node $C$ and the incoming gradients $grad_{in}$ for next node(s) in line 7. If the neighboring node $W$ of $C$ is not visited and it receives all incoming gradient flows, we push $W$ onto the  stack and mark it as visited.

\paragraph{Automatic Differentiation Acceleration and Computation Graph Simplification}
In SHAP when the reference data is fed to the model, there are redundant operators during Shapley value's generation. 
For example, the target point's output is recomputed every time when comparing with a reference point, thus making SHAP inefficient.
Meanwhile, the automatic differentiation's acceleration algorithm inside ONNXExplainer optimizes the existing computing approach and simplifies the computation graph to generate Shapley values. 
Some of the optimization and simplification strategies are like caching commonly-used intermediate outputs during the forward pass for backward propagation.
Figure~\ref{fig:demo_forward} shows a demo neural network graph to explain the details of our optimization and simplification algorithm.
This simple forward symbolic graph contains only two nodes: \textit{MatMul} with weight W and \textit{Sigmoid}, their corresponding dimensions listed in Figure~\ref{fig:demo_forward}.
Figure~\ref{fig:opt_shap} shows two diagrams of how our approach and SHAP compute Shapley values for the NN model in Figure~\ref{fig:demo_forward}.
SHAP explains the input based on an iterative comparison to the sample inside the reference data set one by one. 
In the example in Figure~\ref{fig:opt_shap}, we have 5 reference samples as $R$ of $5\times 32$ to explain one example $X$ of $1\times 32$. 
We summarise how our optimization approach simplifies the computation graph of Shapley values compared to the usual SHAP in two respects:

\begin{itemize}
    \item Forward pass: 
    For SHAP, shown in Sub-figure~\ref{fig:shap_diagram} as the blue blocks,  each sample inside the reference set has to inference both the target and the reference to get the final output difference. 
    Thus for this demo, SHAP has to infer 10 times in total:  5 for the one-time reference 5 for the target input.
    On the other hand, as shown in Sub-figure~\ref{fig:opt_diagram} as the mirror forward pass, the forward symbolic graph in the optimized approach only infers the target once then caches the output and broadcast for further usage.
    $h_{R}$ is the outcome of \textit{MatMul} and also is the input to \textit{Sigmoid} for the reference data. 
    $\hat{y}_{R}$ is the outcome of \textit{Sigmoid} for the reference points. 
    Both $h_{R}$ and $\hat{y}_{R}$ are ingested in computing the multiplier for \textit{Sigmoid} in explaining new data on-the-fly.
    \item Backward pass: 
    This example has one linear operator, \textit{MatMul}, and one nonlinear operator, \textit{Sigmoid}, and their optimized gradient computations are indicated respectively as following.
    \begin{enumerate}
        \item \textit{Sigmoid}:
        As shown in the shaded areas of Figure~\ref{fig:shap_diagram}, since SHAP computes the target output 5 times in the previous forward step, it needs to first split the data then do the subtraction and finally tile back in order to pass  further operators.
        Meanwhile, as the shaded areas in Figure~\ref{fig:opt_diagram} show, the target's output from the forward pass is directly broadcast to the subtraction operator.
        Thus, due to the redundant computation in the forward pass, SHAP has two more \textit{Split}s and one more \textit{Tile} than our optimization approach. 
        Moreover, because the subtraction outputs' dimension is reduced from $10 \times 1$ to $5 \times 1$, the \textit{Greater} in the optimization approach takes half the number of floating-point operations as SHAP. Both SHAP and our approach use the reference's \textit{Sigmoid} gradients as the incoming gradients for \textit{MatMul}. The difference is our approach computes the gradients only one time when constructing the backward symbolic graph and SHAP keeps recomputing the same gradients at run-time.
        
        \item \textit{MatMul}: Similar to the \textit{Greater} operation in \textit{Sigmoid}, the transposed $W$ in the optimization approach multiplies with half the number of floating-point operations that SHAP needs to acquire the gradients.
    \end{enumerate}
\end{itemize}

\subsection{One-shot Deployment}
As shown in Figure~\ref{fig:onnx_exp_dia}, one of the major contributions to the proposed ONNXExplainer is its ability to save the forward neural network and its corresponding graph to calculate Shapley values together in a single ONNX file for one-shot deployment. 
Meanwhile the current open-sourced libraries, such as SHAP~\cite{lundberg2017unified}, need to call their own APIs (Application Programming Interfaces) in order to generate Shapley values, accounting for an extra step during deployment.
Moreover, these libraries depend on other deep learning frameworks, such as TensorFlow or PyTorch, as their computation backend, which makes the on-device deployment even more complicated.

\section{Results}
\label{sec:results}
\subsection{Experimental Settings}
\paragraph{Dataset} We use images of $3\times 224\times 224$ with 10 classes on the ILSVRC-2012 dataset~\cite{deng2009imagenet} to evaluate the explanation time between our optimization approach and SHAP. We use a reference input of all zeros both for the optimization approach and SHAP.

\paragraph{Neural Networks} We use four representative models: VGG19 (V19)~\cite{simonyan2014very}, ResNet50 (R50)~\cite{he2016deep}, DenseNet201 (D201)~\cite{huang2017densely}, and EfficientNetB0 (EB0)~\cite{tan2019efficientnet} in our benchmarks. V19 is one of the neural networks with a large number of weights with 19 weighted layers and no jump layers. R50 is one variant of residual NNs that stack residual blocks, in which skip connections are jump layers to convert regular networks to residual networks via addition. R50 has 107 weighted layers and 16 "add" jump layers. D201 is the largest DenseNet variant which concatenates all previous layers with the current layer to form skip connections. It has 33 jump layers. EB0 is a scaling NN with much fewer parameters and faster speed that is used for on-device platforms. It has 25 jump layers of 9 additions and 16 multiplications\footnote{Details of the NNs and theoretical analysis of computational complexity and memory consumption to explain them can be found in the Appendix C.}.

\paragraph{Machine} We use two machines to perform our benchmarks\footnote{All brand names, logos and/or trademarks are the property of their respective owners, are used for identification purposes only, and do not necessarily imply product endorsement or affiliation by the authors and their associated employers or institutions.}.

Machine A: CPU: AMD EPYC 7513 32-Core Processor; Total Memory:1056 GB

Machine B: GPU: V100; CPU: Intel(R) Xeon(R) Gold 6130 CPU; Total Memory: 790 GB; Nvidia Driver: 510.108.03; CUDA version: 10.1; 


All benchmarks running on three machines use the same library dependencies: \texttt{TensorFlow 1.15.0}, \texttt{onnxruntime-gpu 1.12.0}, \texttt{onnx 1.12.0}, \texttt{Torch 1.13.1}.

\subsection{The Visualization of Explanations from SHAP and optimized ONNX Explainer}
\begin{figure}[ht]
\centering
\includegraphics[width=.99\linewidth]{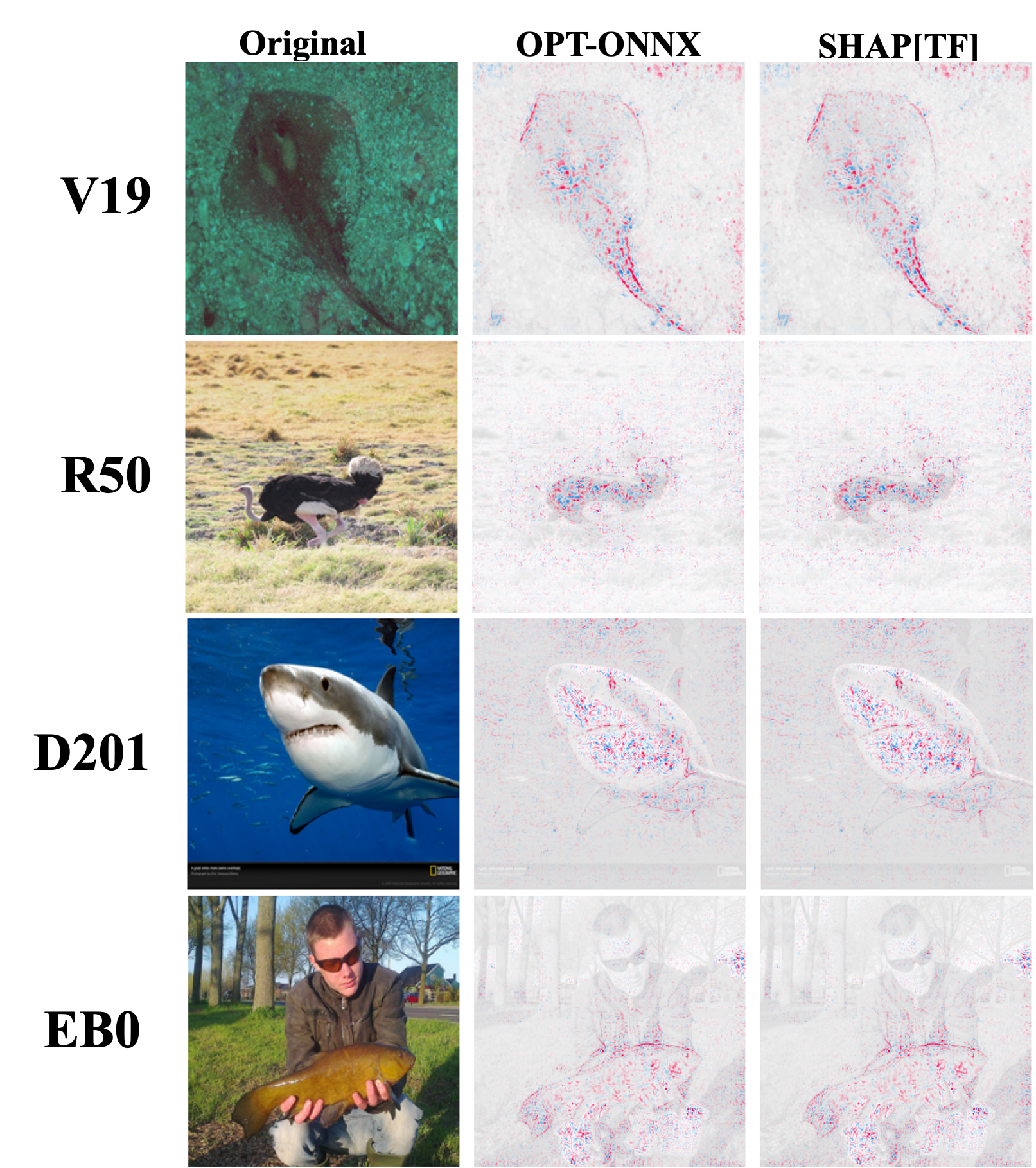}
\caption{Visualization of original images and its corresponding Shapley values simulation images by OPT-ONNX and SHAP (NON-TF). Each row represents a respective model. The first column is the images being explained, the second column is the simulation images by OPT-ONNX, and the last one is simulation images by SHAP[TF].}\label{fig:sv_vis}
\end{figure}

\begin{table}[]
\centering
\begin{tabular}{|c|c|c|c|}
\hline
NN & agreement (\%) & ONNX (\%) &SHAP (\%) \\ \hline
V19 & 100.0 & 0.0 & 0.0 \\ \hline
R50 & 99.4 & 0.6 & 0.0 \\ \hline
D201 &99.8  & 0.2 & 0.0 \\ \hline
EB0 &99.6  & 0.4 & 0.0 \\ \hline
\end{tabular}
\caption{User agreement on closeness of contribution scores between OPT-ONNX and NON-TF. The third column means ONNX is better and the last column means SHAP is better.}
\label{tab:agr}
\end{table}

To evaluate contribution scores obtained by different explainers, we design the following task: in test dataset, we randomly select 10 images from 10 classes to conduct a user study. We do this by using OPT-ONNX and SHAP (NON-TF) to compute Shapley values to explain 100 images, respectively. Then, we only plot simulation images of Shapley values for the predicted class by four models. We arrange original images and simulation images of Shapley values (randomized order) in a row and send 100 rows of images for each model to users. Table~\ref{tab:agr} shows users' agreement on whether the contribution acquired by OPT-ONNX and NON-TF are the same. We observe that all users have a high agreement score (over 99\%) that OPT-ONNX and NON-TF explain the images equally well. In some cases, OPT-ONNX does better than NON-TF. Figure~\ref{fig:sv_vis} shows example images and corresponding contribution score simulation images by OPT-ONNX and NON-TF for each model. It can be concluded that it is hard to tell the difference between the two simulation images visually\footnote{Numerical comparisons between the explainers can be found in the Appendix D.}.

\subsection{Memory Consumption and Latency Analysis}

In this subsection, we use the optimization approach and SHAP described in previous sections for a detailed memory consumption and latency analysis. SHAP supports explaining NNs implemented in TensorFlow (NON-TF)~\cite{abadi2016tensorflow} and PyTorch (NON-PT)~\cite{paszke2019pytorch}. Other than ONNXExplainer (OPT-ONNX), we implement ONNXExplainer with no optimization (NON-ONNX) and optimized SHAPs. In this manuscript, we implement the same optimization process (optimized SHAPs) in TensorFlow (OPT-TF) and PyTorch (OPT-PT) as in ONNX. Then we can compare the latency for three pairs of explainers in the same frameworks: OPT-ONNX versus NON-ONNX, OPT-TF versus NON-TF, OPT-PT and NON-PT. In addition, we use half-precision for the four NN models with GPUs in benchmarks. 

\paragraph{Memory Consumption}
\begin{table}[ht]
\centering
\begin{tabular}{|r|c|c|c|c|}
\hline
 & V19 & R50 & D201 & EB0 \\ \hline
 OPT-ONNX& \textbf{86}/\textbf{166} & \textbf{182}/\textbf{362} & \textbf{78}/\textbf{158} & \textbf{166}/\textbf{255} \\ \hline
 NON-ONNX& 61/121 & 130/256 & 47/93 & 68/135 \\ \hline
 OPT-TF& \textbf{79}/149 & \textbf{157}/\textbf{242} & \textbf{60}/\textbf{115} & \textbf{154}/\textbf{232} \\ \hline
 NON-TF& 76/\textbf{150} & 81/150 & 34/66 & 78/141 \\ \hline
 OPT-PT&\textbf{97}/\textbf{175} &\textbf{112}/\textbf{253}  &\textbf{72}/\textbf{127} & \textbf{114}/\textbf{266} \\ \hline
 NON-PT& 72/163 &107/243  &49/104 & 89/204 \\ \hline
\end{tabular}
\caption{The largest number of reference images that can be used on a V100 GPU [FP32/FP16].}
\label{tab:memory}
\end{table}

As mentioned earlier, the optimization reduces the memory consumption in the forward pass and theoretically in the backward pass too. Non-optimized explainers explain the data one by one and so do its counterparts in the benchmarks (more details in Appendix C). The largest number of reference images is used to measure the memory usage in both optimized and non-optimized explainers. Table~\ref{tab:memory} shows the largest number of reference images that can be used on a V100 GPU for each explainer. In general, the optimized approach can use many more reference images than its counterparts both using single and half precision floating point except for OPT-TF for V19 in half precision. The more reference samples, the more accurate the explanation is. In terms of frameworks, OPT-ONNX for EB0 and V19 and OPT-TF for R50 and D201 gain a superior edge over the number of reference images both in single and half precision. In terms of models, the optimized explainers for EB0 can use more reference images than their counterparts in other three models except for OPT-PT for D201 in single precision and OPT-TF for D201 in half precision. 


\paragraph{Latency Analysis} We infer and explain 100 images in each benchmark. After loading the models, the first few inference requests can be significantly slower at run-time due to deferred initialization and optimizations. Thus we consider the time to explain the first image out of 100 as the cold start time (or warmup time) and average the per-image latency of the rest for each benchmark.

\begin{figure}[ht]
        \centering
        \includegraphics[width=.99\linewidth]{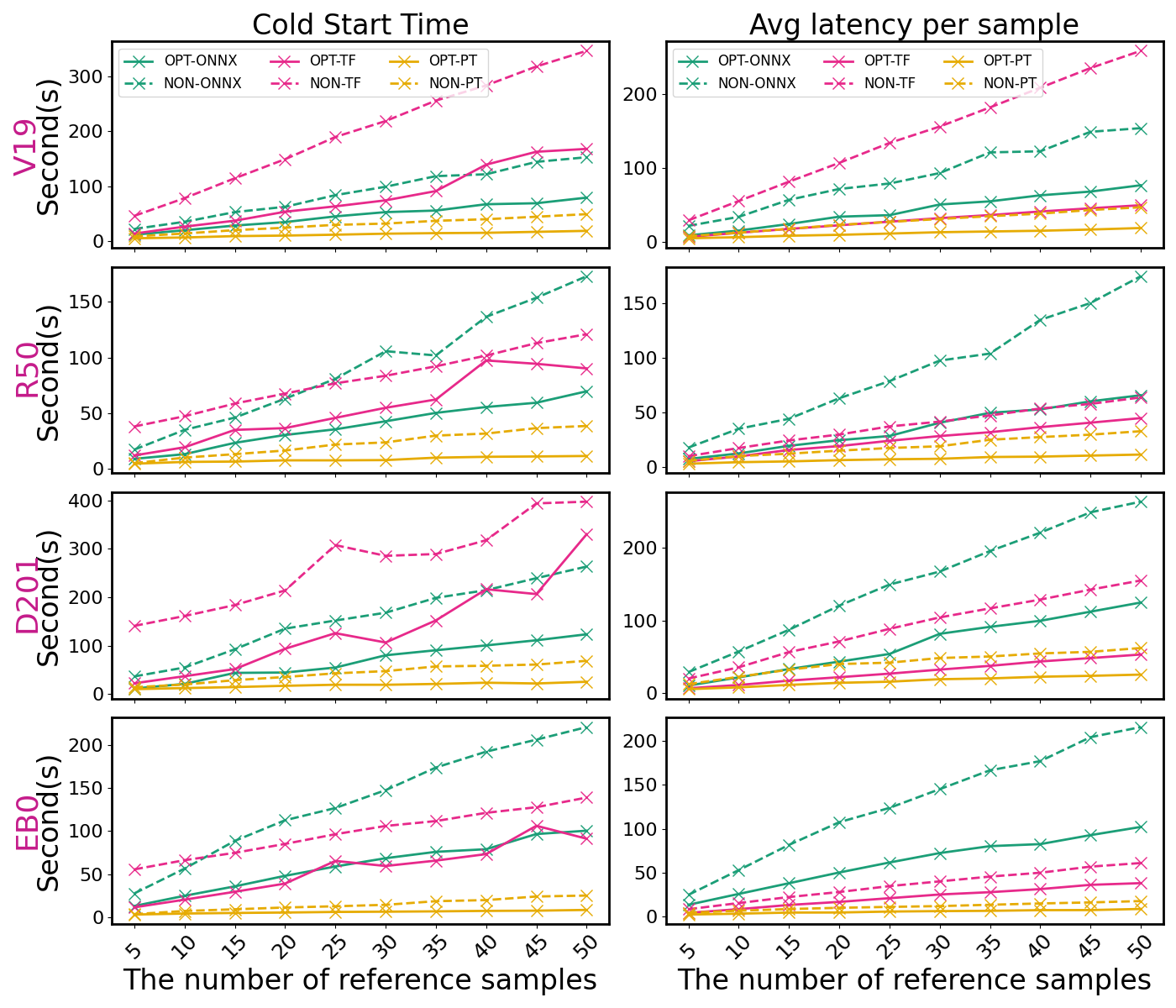}
\caption{Latency Comparison[FP32] between Optimization and Non-optimization/SHAP in three Frameworks (ONNX, TensorFlow, and PyTorch) using CPU cores on Machine A. 
}\label{fig:cpu}
\centering
\end{figure}

We run benchmarks using CPUs on Machine A and the results are shown in Figure~\ref{fig:cpu}. The optimized explainers are much faster at explaining the images than the non-optimized explainers in all settings. Additionally, the speedup efficiency obtained by ONNX and TensorFlow is stronger than PyTorch in all models. Framework-wise and for optimized explainers, PyTorch is the fastest for all models and TensorFlow comes next fastest. 

\begin{figure}[ht]
\centering
        \includegraphics[width=.99\linewidth]{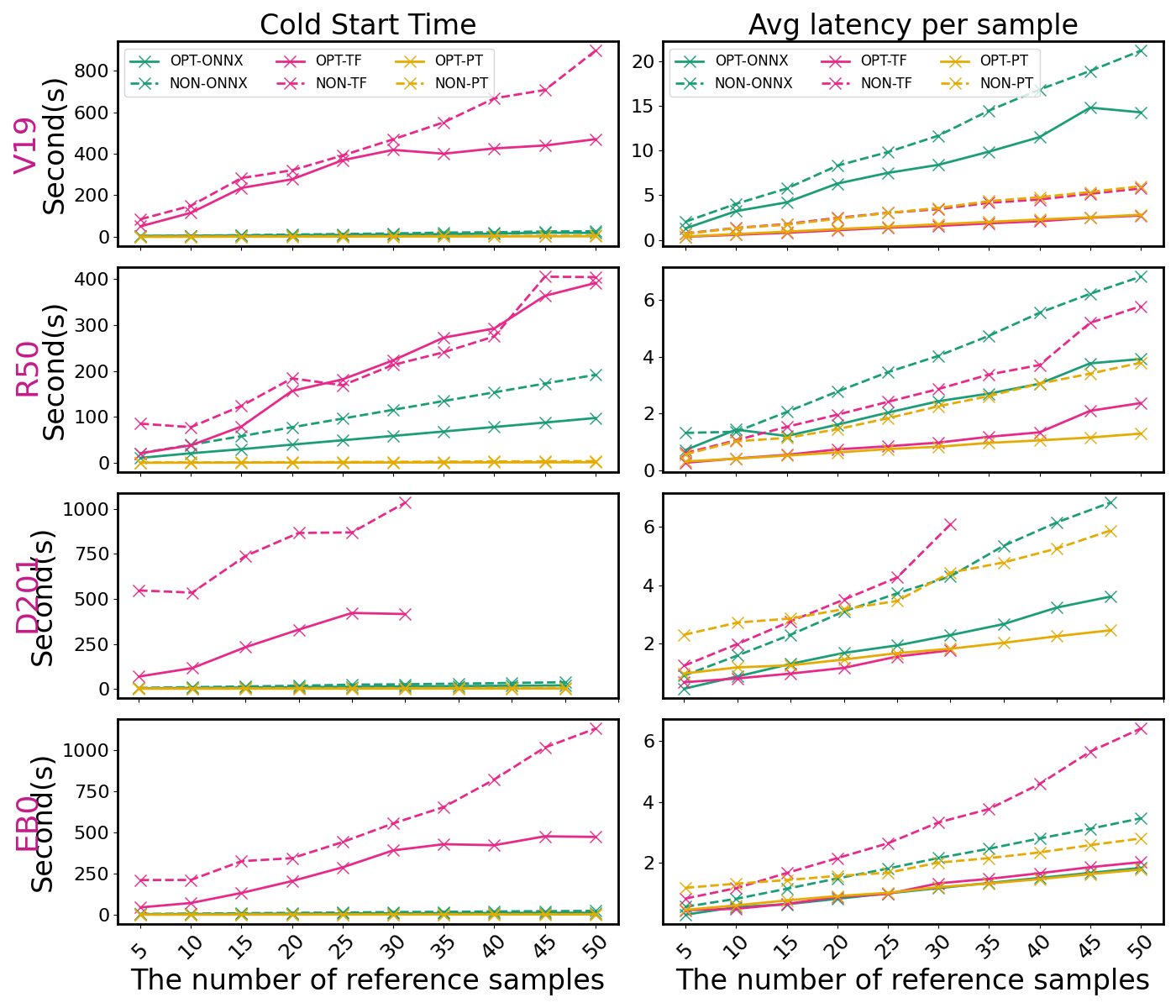}
\caption{Latency Comparison[FP32] between Optimization and Non-optimization/SHAP in three Frameworks (ONNX, TensorFlow, and PyTorch) using GPU on Machine B. We use 45, 30, and 45 for NON-ONNX, NON-TF, and NON-PT for DenseNet201 because of memory limitation.
}\label{fig:fp32}

\centering
\end{figure}

Figures~\ref{fig:fp32} show latency comparisons on a V100 GPU\footnote{Half-precision benchmarks can be found in the Appendix E.}. It can be observed that optimized explainers are superior to non-optimized explainers in most of benchmarking configurations with one exception of cold start time for R50 by TensorFlow. In this exception, OPT-TF spends a little more time explaining R50 than NON-TF during warmup time in a few data points. Compared to benchmarks on CPU, TensorFlow and ONNX both need to significantly warm up the models. In particular, it takes TensorFlow hundreds of seconds to explain the first image in all models. This tells us substantially warming up the models would guarantee smooth production traffic without any delays. In terms of average latency, OPT-TF and OPT-PT performs very close in all models. It seems that OPT-TF has the largest acceleration gains than its peers for R50, D201, and EB0 and OPT-ONNX surpasses OPT-TF and OPT-PT in speeding up the explanation for V19.

\section{Conclusion}
\label{sec:conclusion}
In this work we propose the ONNXExplainer to explain NNs using Shapley values in the ONNX ecosystem. In ONNXExplainer, we build our own automatic differentiation which enables one-shot deployment of NNs in inference pipelines, such as Triton and ONNXruntime. 
The optimization by precomputing outcomes from the reference data reduces lots of redundant computations in explaining NNs. 
We develop and benchmark optimized explainers and non-optimized explainers in three major deep learning frameworks (ONNX, TensorFlow, and PyTorch) and test and compare the explainers using four typical neural networks: VGG19, ResNet50, DenseNet201, and EfficientNetB0. 
The benchmarks show that our optimized explainers significantly outperform counterparts in terms of inference resources usage and explanation latency.

Future work will continue to generalize ONNXExplainer by: 1) supporting gradients/multiplier computation for more operations, such as \textit{Loop}, \textit{GatherND}; 2) supporting more NN structures, such as Bidirectional LSTM/GRU models; 3) further optimization of the gradients computation graph.

\section{Author Contributions}
Conceptualization, methodology, Algorithm, Y.Z.; Experiments, Y.Z., R.H, N.K.; writing–original draft preparation, Y.Z., R.H, N.K., C.L.; writing–review and editing, R.H, Y.Z., N.K., S.A., C.L., C.C., Y.G.; visualization, Y.Z., R.H.

\bibliography{aaai24}

\clearpage
\begin{appendices}
\setcounter{figure}{0}
\setcounter{table}{0}
\setcounter{equation}{0}
\setcounter{secnumdepth}{1}

\section{Parser code}
\begin{listing*}[htb]%
\caption{Network Neural parser code and build backward pass graph}%
\label{lst:parser}%
\begin{lstlisting}[language=Python]
from collections import defaultdict
class GraphVertex: #define vertex of backward graph
  def __init__(self, 
    node=None, #computation node
    neighbors=None, #neighbors in backward pass
    flowin_grads=None, #incoming gradients
    flowout_grad=None, #outgoing gradients
    forward_times=1, #num of flowing in grads
    pass_grads=None
  ):
  self.node = node
  self.neighbors = neighbors if neighbors is not None else []
  self.flowin_grads = flowin_grads if flowin_grads is not None else []
  self.flowout_grad = flowout_grad
  self.forward_times = forward_times
  self.pass_grads = pass_grads if pass_grads is not None else {}

nodes = get_nodes() #get computation nodes from ONNX model
input2node = get_input2node() #get mapping from differential inputs to computation node
output2node = get_output2node()#get mapping from outputs to computation node
backward_graph = defaultdict(GraphVertex)
backward_graph['backward_start'].neighbors.append(nodes[0])

for node in nodes:
  for iname in node.input:
    if iname in output2node:
      curr_nodes = input2node[iname]
      next_node = output2node[iname]
      for curr_node in curr_nodes:
        for output in curr_node.output:
          if output not in backward_graph:
            backward_graph[output].node = curr_node
            if next_node not in backward_graph[output].neighbors:
              backward_graph[output].neighbors.append(next_node)
for name in input2node:
  if name in backward_graph:
    if len(input2node[name]) > 1:
      backward_graph[name].forward_times = len(input2node[name])
\end{lstlisting}
\end{listing*}
In Python Code List~\ref{lst:parser}, a backward pass graph is first established to keep the structures of forward pass. The information in each vertex contains the computation node itself, neighbors of the current computation node in the backward graph and the number of neighbors, in- or out-flowing gradients, and the optional argument \texttt{pass\_grads} to tell whether the input to the nodes is differential with the model input as shown in \texttt{GraphVertex} in Python Code~\ref{lst:parser}. Some operators, such as \textit{Mul} and \textit{Add}, allow two inputs to differentiate with the model input and others, such as \textit{MatMul} and \textit{Gemm}, allow up to two inputs to vary when only one input differentiates with the model input. We need to determine whether the out-flowing  gradients of the current computation node can be passed to its neighbors via \texttt{pass\_grads} when building the backward graph using Neural Network Parser. The meat of the algorithm for Parser is shown in Python Code List~\ref{lst:parser}. 
Function \texttt{get\_input2node} establishes a mapping between the differential inputs of a computation node and itself. 
The argument \texttt{pass\_grads} retains the information on whether the input of a computation node is differential with the model input for some operators, such as \textit{Mul}. 
ONNXExplainer only passes gradients to the succeeding computation node with the differential inputs in the backward pass.

\section{Gradients/Multipliers Computation}
In this section, we provide gradients/multipliers computation for several example operators.
\begin{itemize}
    \item \textit{Concat}: The concatenation is a linear operator, which has no local gradient. We only need to split and pass the incoming gradient to the successors in the backward pass according to the portion of how the inputs to \textit{Concat} are concatenated in the forward pass.
    
    \item \textit{Mul}: The effect of multiplication can be either nonlinear or linear, depending on whether both inputs to the multiplication operation are differentiable w.r.t the model input. If so, we use  \textit{RevealCancel Rule}~\cite{shrikumar2017learning} to compute adjusted gradients. Otherwise, we multiply the incoming gradients with the input to the multiplication that is not differentiable w.r.t the model input to compute outgoing gradients. In case of broadcasting, the smaller input sums the incoming gradients over the axes that this input is "broadcast" across with the larger input of multiplication.
    
    \item \textit{Matmul}: The matrix multiplication is a linear operator. Its local gradient is the transpose of the weight w.r.t. the input. Multiplying its local gradient with incoming gradient gives the outgoing gradient to the successors in the backward pass.
    \item \textit{Conv}: Convolution is a linear operator. We can use \textit{Conv} to compute outgoing gradients for the \textit{Conv}. Let's take 2D convolution as an example. Consider a case where the input is $4\times 4$, the filter is $2\times 2$, and the stride is $1\times 1$:
    \begin{align}
    input&=\begin{bmatrix}
        X_{00}&X_{01}&X_{02}&X_{03}\\
        X_{10}&X_{11}&X_{12}&X_{13}\\
        X_{20}&X_{21}&X_{22}&X_{23}\\
        X_{30}&X_{31}&X_{32}&X_{33}
    \end{bmatrix}\\
    filter&=\begin{bmatrix}
        F_{00}&F_{01}\\
        F_{10}&F_{11}
    \end{bmatrix},
    \end{align}
    where $X_{ij}$ is $batch\times C_{in}$ and $F_{ij}$ is $C_{out}\times C_{in}$. With valid padding, the output is $batch\times C_{out}\times 3\times 3$. In back-propagation, the incoming gradient also should be $batch\times C_{out}\times 3\times 3$. The computation of outgoing gradient for a 2D \textit{Conv} is to convolute the padded incoming gradient $grad_{in}$ using the transposed and flipped filter $filter^G$ as follows:
    \begin{align}
        grad_{in}&=\begin{bmatrix}
            0&0&0&0&0\\
            0&G^\prime_{00}&G^\prime_{01}&G^\prime_{02}&0\\
            0&G^\prime_{10}&G^\prime_{11}&G^\prime_{12}&0\\
            0&G^\prime_{20}&G^\prime_{21}&G^\prime_{22}&0\\
            0&0&0&0&0
        \end{bmatrix}\\
        filter^G&=\begin{bmatrix}
            F_{11}^{T}&F_{10}^{T}\\
            F_{01}^{T}&F_{00}^{T}
        \end{bmatrix}
    \end{align}
    With strides more than one, plus zero-padding the incoming gradient, dilations are used to control the spacing in the incoming gradient.

    \item \textit{Sigmoid}: The Sigmoid is a nonlinear operation. The computation of adjusted gradients is denoted as following:
    \begin{align}
        \sigma(x) &= \frac{1}{1+e^{-x}}\\
         grad^* &=
            \begin{dcases}
            \sigma(x)(1-\sigma(x)) & \text{if } x - r < 1e-6 \\
            \frac{\sigma(x) - \sigma(r)}{x - r} & \text{otherwise}
            \end{dcases}
    \end{align}
Where $\sigma(x)$ is the output of \textit{Sigmoid}, $x$ is the input to some neurons for the data being explained, and $r$ is the input to some neurons for reference data. If $x - r < 1e-6$ returns true, we use the original gradients of \textit{Sigmoid}. Otherwise, we use the multiplier for \textit{Sigmoid} by the  \textit{Rescale Rule}~\cite{shrikumar2017learning}. Multiplying $grad^*$ with incoming gradients can give us the outgoing gradients. Note that most of activation functions use the same manner to obtain $grad^*$ except for \textit{Softmax} which uses \textit{RevealCancel Rule}~\cite{shrikumar2017learning}.

    \item \textit{Maxpooling}: Maxpooling is a nonlinear operation. We use the following to define the adjusted gradients for it~\cite{lundberg2017unified}.
    \begin{align}
        C &= \max(y_{x}, y_{r})\\
        M_x &= (C-y_{r})\times grad_{in} \\
        M_r &= (y_{x}-C)\times grad_{in} \\
        grad^*_{out} &=
            \begin{dcases}
            0s & \text{if } x - r < 1e-7 \\
            \frac{\frac{\partial M_x}{\partial x} + \frac{\partial M_r}{\partial r}}{x - r} & \text{otherwise}
            \end{dcases}
    \end{align}
    Where $x$ and $r$ are the inputs to maxpooling neurons for the data and reference data, respectively, and $y_x$ and $y_r$ are outputs of these neurons. $C$ is the cross maximum between $y_x$ and $y_r$ element-wise. We multiply incoming gradients $grad_{in}$ with $C-y_{r}$ to attain cross positioned incoming gradients $M_{x}$. Likewise, we have $M_{r}$. If $x - r$ is less than $1e-7$,  the outgoing gradients $grad^*_{out}$ are zeros. Otherwise, we use the sum of positioned gradients of \textit{Maxpooling} w.r.t $x$ and $r$ divided by $x - r$ as outgoing gradients $grad^*_{out}$. The incoming gradient is only passed back to neurons that achieve the maximum and all other neurons have zero gradients when calculating gradients for maxpooling operations. Note that the gradients accumulate if the same neurons achieve the maximum in different pooling windows. \textit{GlobalMaxPooling} is a special case of \textit{Maxpooling} whose pooling window size is the same as the input spatial.
    
    On the other hand, \textit{Avgpooling} is a linear operation. To compute gradients w.r.t. input of an \textit{Avgpooling} operation, the incoming gradients are distributed equally to the the locations within the pooling window. The gradients can accumulate if two pooling windows overlap. Similarly, \textit{GlobalAvgPooling} is a special case of \textit{Avgpooling} whose pooling window size is the same as the input spatial.

\end{itemize}

\begin{table*}[ht]
\centering
\begin{tabular}{|c|c|c|c|c|}
\hline
 Comparison& VGG19 & ResNet50 & DenseNet201 & EfficientNetB0 \\ \hline
 OPT-ONNX versus SHAP[NON-TF]& 99.5\% & 96.6\% & 96.7\% & 82.6\% \\ \hline
 OPT-TF versus SHAP[NON-TF] & 99.2\% & 99.2\% & 99.0\% & 87.9\% \\ \hline
 OPT-ONNX versus NON-ONNX& 100\% & 98.7\% & 100\% & 99.0\% \\ \hline
\end{tabular}
\caption{Numerical comparison of Shapley values between two different explainers. We use $\lvert a-b\rvert < (atol + rtol \times \lvert b \rvert)$ to compute the percentage of closeness of two sets of Shapley values, where $a$ and $b$ are two sets of Shapley values, $atol$ is the absolute tolerance parameter, and $rtol$ is the relative tolerance parameter. Here, $atol$ is set to $1.0e-8$ and $rtol$ is set to $1.0e-5$.}
\label{tab:num_sv}
\end{table*}

\section{Neural Networks Used in Benchmarks and Theoretical Analysis of Computational Complexity and Memory Consumption in Explanation}
\begin{table*}[ht]
\centering
\begin{tabular}{|c|c|c|c|c|}
\hline
NN & \#para & depth & $\frac{\#para}{depth}$ & jump type \\ \hline
VGG19(V19) & 143.7M & 19 & 7.56 & - \\ \hline
ResNet50(R50) & 25.6M & 107 & 0.24  & add \\ \hline
DenseNet201(D201) & 20.2M & 402 & 0.05  & concat \\ \hline
EfficientNetB0(EB0) & 5.3M & 132 & 0.04  & add \& mul \\ \hline
\end{tabular}
\caption{Four representative models. Depth counts the number of layers with parameters.}
\label{tab:backbone}
\end{table*}
Table~\ref{tab:backbone} shows the four Neural Networks used in the benchmarking experiments. The ratio $\frac{\#para}{depth}$ is the rate between \#para, the total number of parameters, and depth, which can be considered as a indicator of parallelizability of the neural network. The larger width, the easier for the GPUs to parallelize to compute neural networks. From V19 to EB0 in Table~\ref{tab:backbone}, the ratio becomes smaller and smaller. So is $\#para$. Now let us discuss the hypothetical memory consumption and computational complexity of the four NNs in explaining data.

\paragraph{Memory Consumption} The memory consumption decreases by a factor of $2B$ in the forward pass of the optimized approach because the optimized approach only consumes the image being explained.
Comparatively, SHAP consumes $2B$ images in the forward pass.
Theoretically, both methods need similar amounts of memory in the backward pass. The difference is SHAP computes the intermediate outputs for the reference images at run-time, but the optimized approach precomputes them for the reference images when building the backward symbolic graph. In practice, our implementations of backward pass might need more memory than SHAP because users have no access to the intermediate outputs of forward pass which usually are used to compute gradients in training~\cite{abadi2016tensorflow, paszke2017automatic, paszke2019pytorch}.

\begin{figure}[ht]
\centering
\includegraphics[width=.99\linewidth]{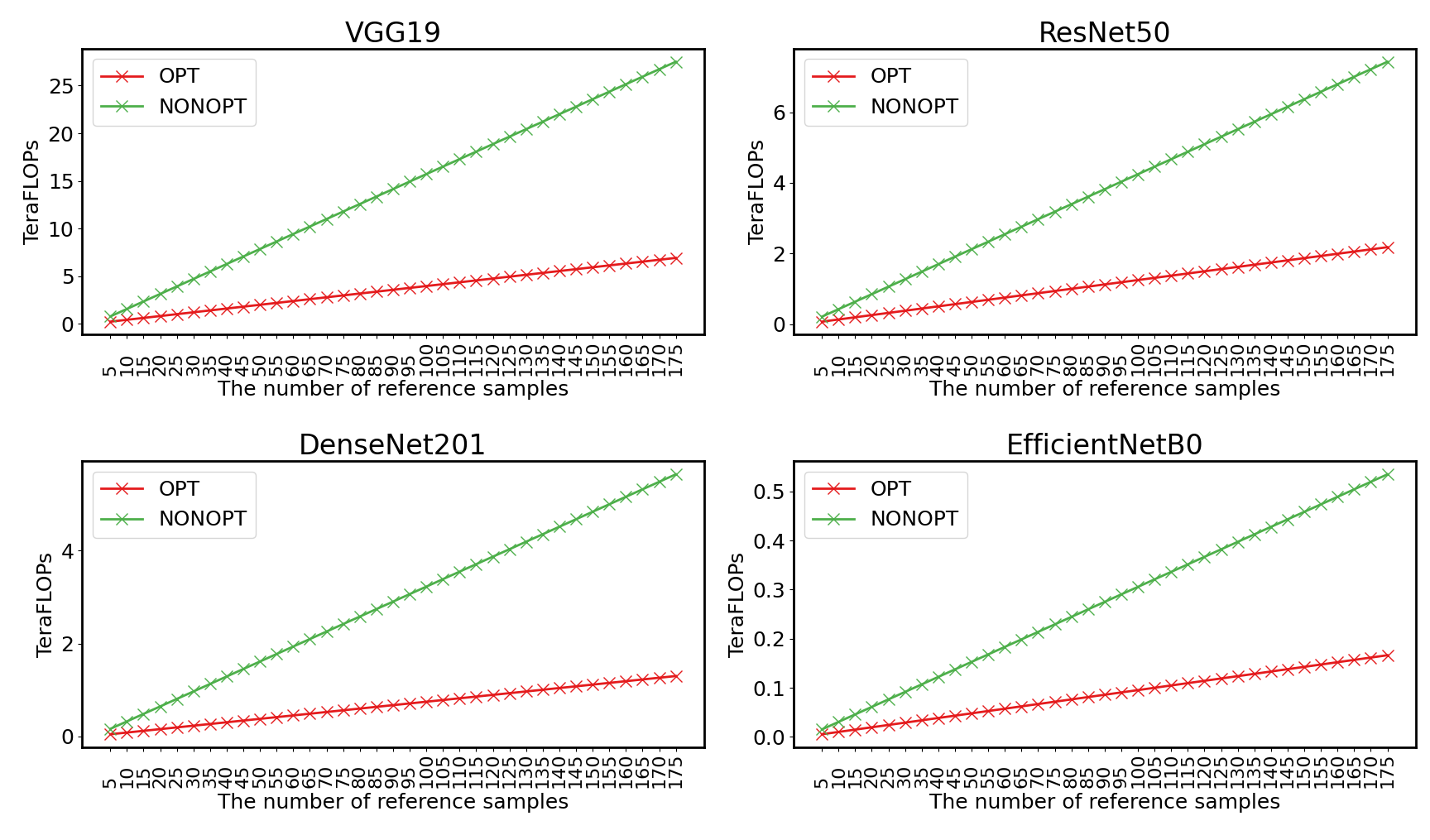}
\caption{The estimated computational complexity of backbone networks in terms of reference image size ($B$) per explanation image by counting TeraFLOPs manually. The red lines represent the complexity using our optimization approach and the green lines represent it using SHAP(non-optimization approach).}\label{fig:flops}
\end{figure}

\paragraph{Estimated Computational Complexity} Four backward symbolic graphs are built for V19, R50, D201, and EB0, respectively. FLOP stands for floating point operation, which can be used to measure the count of floating point operations of a model. A TeraFLOP means one trillion ($10^{12}$) floating-point operations. TeraFLOPs are estimated theoretically for both forward and backward symbolic graphs built by SHAP and the optimization method for each neural network shown in Figure~\ref{fig:flops}. It is found that the optimization consistently decreases the complexity over reference image size for all networks. The count of decreased TeraFLOPs to compute Shapley values soars quickly for all networks when increasing reference samples until it reaches a certain number of images.

\section{Numerical Comparison of Shapley Values between Explainers}

We use four explainers (OPT-ONNX, NON-ONNX, OPT-TF, and NON-TF) to compute Shapley values for 500 images in test dataset (Models in ONNX and TensorFlow share same weights in inference.). To verify whether our methods have close Shapley values with SHAP (TensorFlow), we compare Shapley values between optimized and non-optimized explainers numerically. The results are shown in Table~\ref{tab:num_sv}. Overall, the Shapley values by optimization method are closely matched with Shapley values by SHAP.

\section{Latency Analysis}

\begin{figure}[ht]
\centering
\includegraphics[width=.99\linewidth]{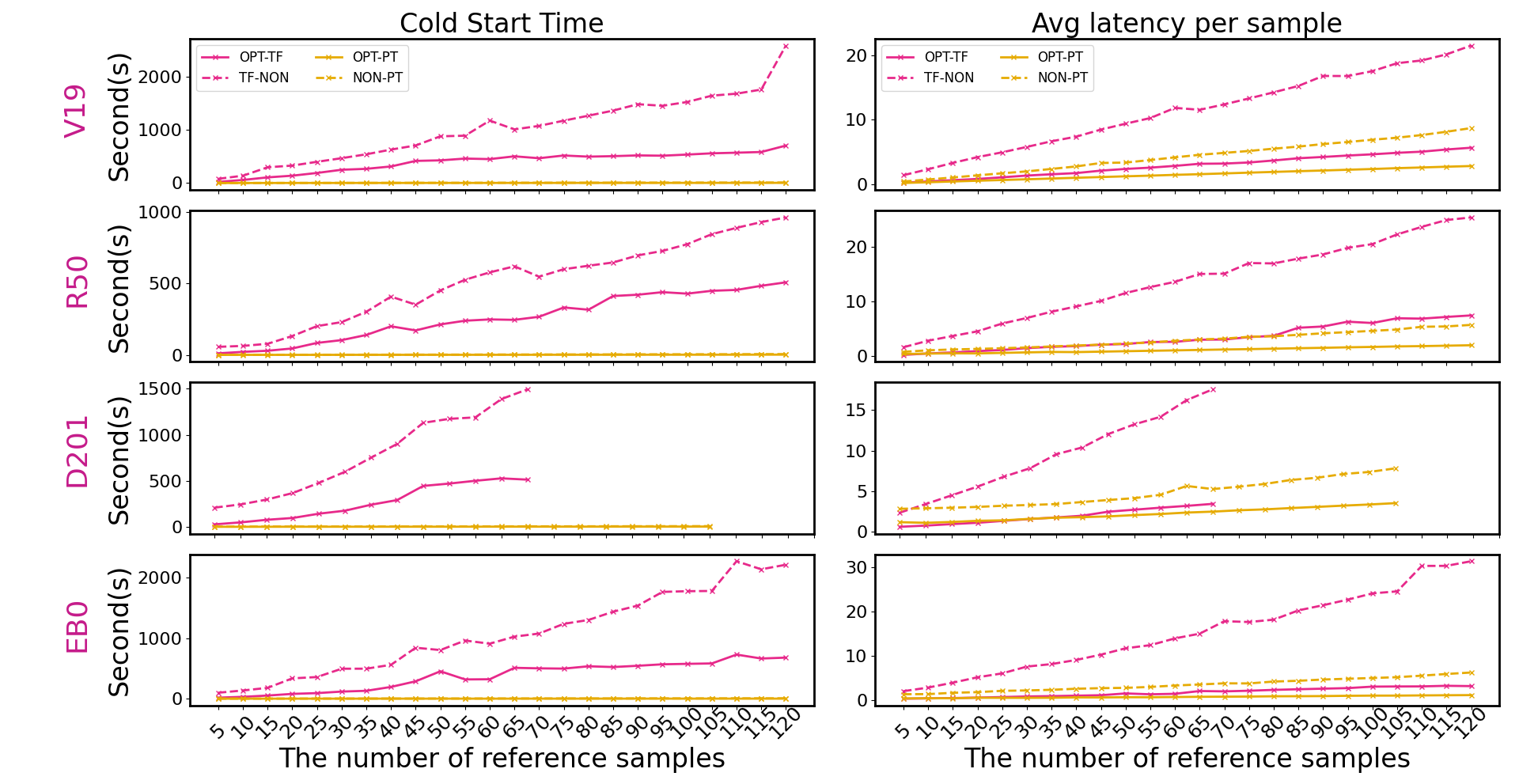}
\caption{Latency Comparison[FP16] between Optimization and Non-optimization/SHAP using two Frameworks (TensorFlow, and PyTorch) on Machine B with GPU. We use 65 and 100 for NON-TF and NON-PT for DenseNet201 because of memory limitation.
}\label{fig:fp16}
\end{figure}

We also run benchmarks using half precision floating-point on one V100 GPU. Figure~\ref{fig:fp16} shows the latency over the number of reference images for four models in two frameworks. TensorFlow models requires a great amount of time to warm up the model. OPT-TF gains much more acceleration than OPT-PT.

\begin{figure}[ht]
\centering
\includegraphics[width=.99\linewidth]{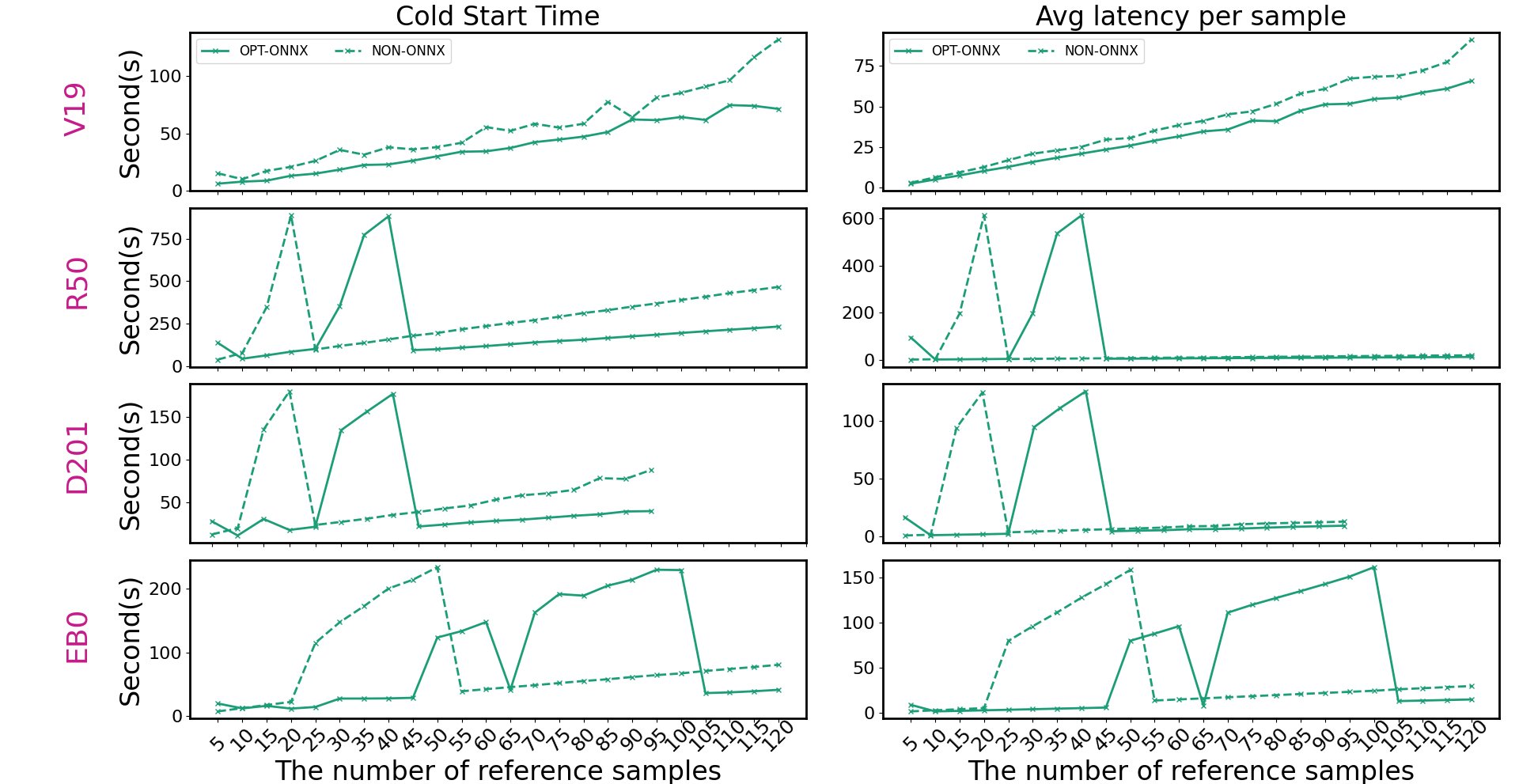}
\caption{Latency Comparison[FP16] between Optimization and Non-optimization ONNXExplainer on Machine B with GPU. We use 90 for D201.
}\label{fig:fp16_spikes}
\end{figure}

\begin{figure}[ht]
\centering
\includegraphics[width=.99\linewidth]{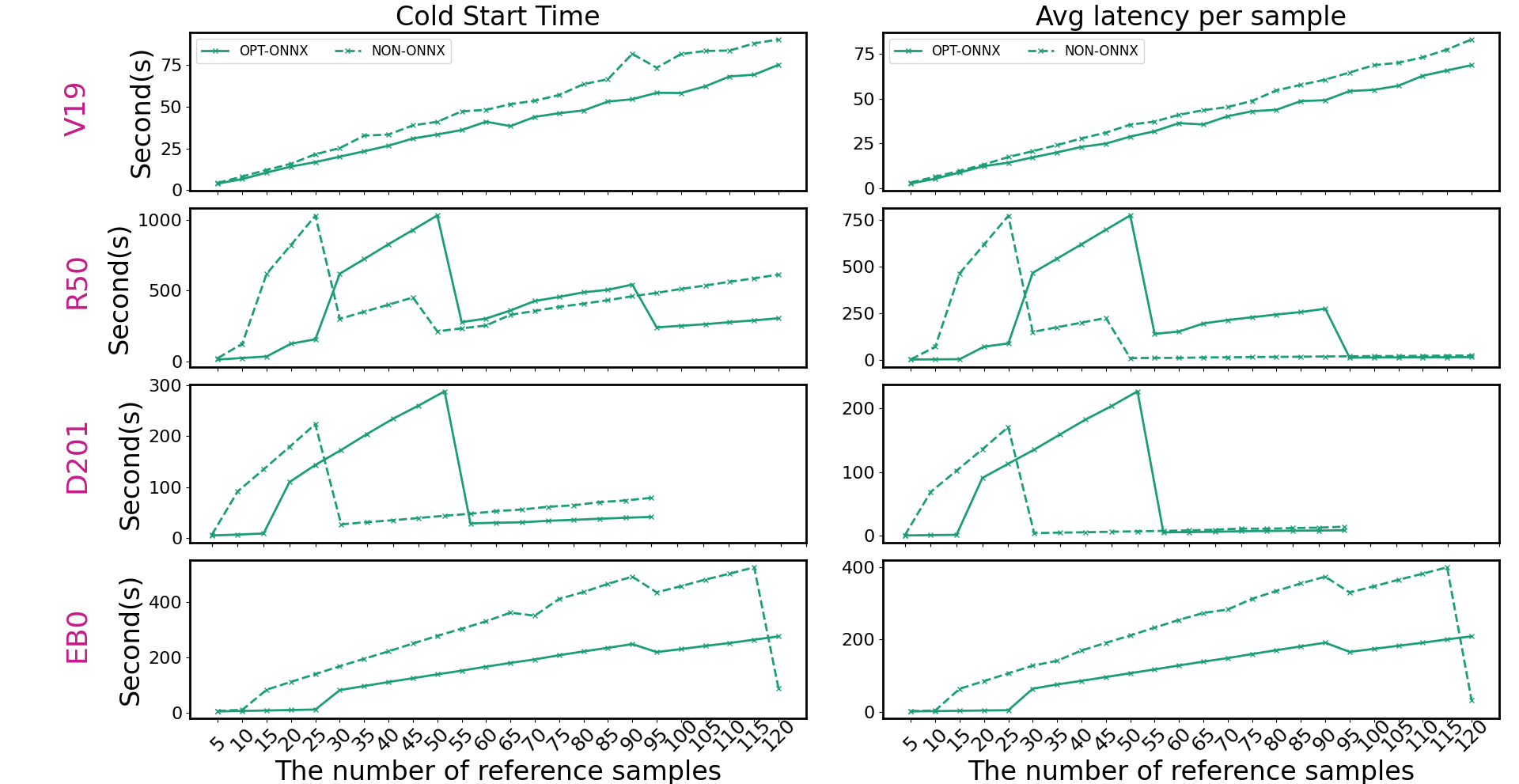}
\caption{Latency Comparison[FP16] between Optimization and Non-optimization ONNXExplainer on a different Machine with V100 GPU. We use 90 for D201.
}\label{fig:fp16_spikes2}
\end{figure}

Figure~\ref{fig:fp16_spikes} shows the latency in ONNX for four models. Surprisingly, there are some spiking latency points both in OPT-ONNX and NOT-ONNX for R50, D201, and EB0. This phenomenon might be a systematic reaction of hardware, software, parallel optimization, and ONNX implementations, which needs future research. Excluding the spiking points, we can find that OPT-ONNX outperforms NON-ONNX in terms of latency. We run the same benchmarking experiments on a different Machine with V100 GPU and the Nvidia Driver (455.23.05) and CUDA version (11.1) are different from Machine B in ONNX ecosystem. The spikes are also found in R50, D201, and EB0 as in Figure~\ref{fig:fp16_spikes2}. In the new machine, the spiking points occur with different input sizes than the machine B. Interestingly, in addition to the increasing number of spikes seen in EB0 in Figure~\ref{fig:fp16_spikes2}, the spikes we observe are frequently lower on optimized ONNXExplainer than non-optimized ONNXExplainer.

\end{appendices}

\end{document}